\pdfoutput=1

\documentclass[11pt]{article}

\usepackage[]{ACL2023}

\usepackage{times}
\usepackage{latexsym}

\usepackage[T1]{fontenc}

\usepackage[utf8]{inputenc}

\usepackage{microtype}

\usepackage{inconsolata}

\usepackage{soul}
\usepackage{graphicx}

\title{Leveraging the power of transformers for guilt detection in text}
\author{Abdul Gafar Manuel Meque,  Jason Angel, Grigori Sidorov, Alexander Gelbukh \\
Instituto Politécnico Nacional (IPN) \\ Centro de Investigación en Computación (CIC) \\ Mexico City, Mexico
  }

\begin{document}
\maketitle
\begin{abstract}
In recent years, language models and deep learning techniques have revolutionized natural language processing tasks, including emotion detection. However, the specific emotion of guilt has received limited attention in this field. In this research, we explore the applicability of three transformer-based language models for detecting guilt in text and compare their performance for general emotion detection and guilt detection. Our proposed model outformed BERT and RoBERTa models by two and one points respectively. 
Additionally, we analyze the challenges in developing accurate guilt-detection models and evaluate our model's effectiveness in detecting related emotions like "shame" through qualitative analysis of results.
\end{abstract}

\section{Introduction}


Guilt is an emotion characterized by the feeling of unease or remorse due to perceived moral transgressions or failure to meet personal standards. According to \citet{rawlings1970a} guilt is particularly complex due to its different manifestations, which can be categorized as reactive, existential, or anticipated guilt, depending on whether it pertains to past, present, or future actions, respectively. 
And although feeling guilty serves as a moral compass, promoting self-reflection, empathy, personal growth, and positive change, when experienced in overabundance, culpability can have adverse effects on one's emotional well-being including stress, anxiety, and depression \cite{tangney2002a}.  Thus, the ability to detect guilt becomes crucial as it paves the way for future endeavors aimed at averting negative outcomes.

From the perspective of Natural Language Processing (NLP), the detection of emotions such as guilt has gained significant popularity in recent years with a notable increment in emotions to be detected, the source of the annotation and the modalities being used \cite{alvarez2021uncovering, garcia2017emotion}. Certainly, the emergence of advanced deep learning techniques like language models \cite{systematic-review-kusal2023}, has facilitated the examination of emotions from a linguistic standpoint, as these models can efficiently process vast amounts of textual data and interpret emotions accurately \cite{acheampong2020text}. However, despite the substantial interest in emotion detection as a whole, there is still a need for further research on specific emotions. Guilt, in particular, has been relatively overshadowed by more popular and general emotions such as sadness or happiness \cite{callcenterbolo2023, angerkim2010}. And to the best of our knowledge, there is only one existing work \cite{meque2023machine} that explores guilt, with classic machine learning.
This study aims to explore the capability of language models in guilt detection. We also introduce GuiltBERT, a masked language model built upon BERT \cite{devlin2018bert}, as an emotion-specific approach to identifying guilt-related linguistic cues and patterns. 
This work makes significant contributions by shedding light on the detection of the less explored emotion of guilt and enabling further research in this domain. Our model achieves comparable results to existing emotion detection models while setting a new state-of-the-art for guilt detection.




The rest of the paper is organized as follows: Section 2 offers a succinct overview of prior research on guilt detection and the different datasets that include guilt as a target emotion for detection. Section 3 examines existing datasets for the study of text-based emotion detection, specifically those that include guilt. Section 4 provides a detailed description of the GuilBERT architecture its training process and the state-of-the-art models used for the evaluation. Section 5 presents the evaluation results of GuilBERT. Section 6 presents a comprehensive analysis of the strengths and limitations of our model regarding the detection of guilt and other emotions. Lastly, in section 7 we conclude by highlighting our contributions and outlining avenues for future research.

\section{Related work}
Guilt has predominantly been studied within the humanities.
Particularly, the literature shows the detection of guilt has been vastly motivated in forensic science because of its applications in criminal investigation and legal proceedings. However, these works focus on detecting deception and determining if a person "is" guilty of a certain prohibited act, rather than examining if the subject "feels" guilty. Our work is centered on the emotional side instead,
in which individuals feel responsible for wrongdoing or violating their own ethical standards because of one's actions or omissions. In this sense, the detection of guilt is closer to the detection of remorse, self-blame, and in some cases, regret and suicide ideation. 
There are few works on this side, some approach the detection using body language expressions \cite{bodyjulle2020} or audio recordings \cite{audioshinde2020} as input. 

Regarding text-based approaches, the ISEAR dataset \cite{ISEAR-Scherer_1994} was probably the first to include guilt as one of the classic four emotions (joy, anger, fear, and sadness), making the guilt emotion more visible and providing researchers with a valuable benchmark to evaluate guilt detection models. 
Guilt was included because is considered to be essentially limited to the human species, like other self-reflexive emotions such as disgust and shame. In this work, the author also found that guilt is quite similar to shame except for longer duration and the absence of marked physiological response symptoms.

Unlike more popular emotions such as happiness and anger \cite{angerkim2010, callcenterbolo2023}, the emotion of guilt has been studied only in a multiclass setting rather than individually, presumably because of its complexity, low resource availability, and minor commercial applications. 
The next subsection presents the publicly available datasets used for training and evaluating guilt detection models, contributing to the advancement of research in this area.


\subsection{Datasets for guilt detection}

Despite the existence of several datasets for emotion detection such as GoEmotions \cite{goemotionsdemszky2020}, TEC \cite{TECmohammad2012}, or Universal Joy \cite{universaljoylamprinidis2021}, very few have considered the guilt emotion in the list of classes and almost no one has dedicated a special focus to it. Here we present a brief description of the existing datasets for guilt detection\footnote{A link to access the original data is publicly available in the cited sources}. 

\begin{enumerate}    

    \item \textbf{VENT:} Vent is the largest existing dataset for emotion detection \cite{ventlykousas2019} it comprises more than 33 million samples self-annotated by nearly 1 Million semi-anonymized users across 705 emotions, among "feeling guilty" is one of them. Most of the samples, (around 93\%) are written in English.

    \item \textbf{ISEAR:} a classic and popular resource for emotion detection \cite{ISEAR-Scherer_1994} created as a result of asking 2921 participants (from 37 countries and 5 continents) to fill a series of questionnaires about their experiences and reactions toward events.
    
    \item \textbf{CEASE:} is a corpus of nearly 205 suicide notes in English comprising 2393 sentences (samples) annotated across 15 labels that include the guilt emotion among them \cite{ceaseghosh2020}. It is worth mentioning that the majority of these samples (58\%) are labeled using the "neutral sentiment tags" of "instruction" and "information", leaving a minor percentage of samples corresponding to emotions (42\%) and particularly very little (3\%) for the guilt emotion.
    
    \item \textbf{VIC:} This dataset leverage other datasets that include the guilt emotion, namely, Vent, ISEAR, and CEASE to create a new dataset focused on guilt detection \cite{meque2023machine}. The result was a combination of fragments of the previous dataset, it comprised 4622 texts, half of which are labeled with guilt and the other half labeled with other 27 different emotions.

\end{enumerate}

\section{Methodology}
In this section, we present our experimentation setting and the procedure we followed to process the datasets being used for the tasks of general emotion detection and guilt detection. 

\subsection{Experimentation setting}
For our experiments, we give high consideration to the work of \citet{adoma_etal_2020} where different Transformer-based models are tested for emotion detection in the ISEAR dataset. The authors finetuned four pre-trained models, namely the Bidirectional Encoder Representations from Transformers (BERT) \cite{devlin2018bert}, the Generalized Autoregression Pre-training for Language Understanding (XLNet), the Robustly optimized BERT pre-training Approach (RoBERTa), and DistilBERT, on ISEAR dataset, first presented by \cite{ISEAR-Scherer_1994}. 

Among the available models, we selected BERT and RoBERTa as the baseline for our experiments, and design GuiltBERT as the intervention model based on the BERT architecture. This selection ensures a fair comparison and meaningful evaluation. Among these, the RoBERTa model has demonstrated superior performance on the emotion detection task and was considered the state-of-the-art model in the ISEAR dataset \cite{adoma_etal_2020}. 



\begin{enumerate}
    \item \textbf{Baseline models:} we used the models of BERT and RoBERTa as the baseline for our experiments and replicated the configuration proposed in \citet{adoma_etal_2020} as their models are considered a state-of-the-art reference for emotion detection. We diligently followed the instructions provided in their report, ensuring the utmost consistency and comparability in our experimental setup. This entails employing identical data preprocessing techniques, hyperparameter tuning, and training and testing configurations as outlined in the original study. After training, we evaluated the performance of each model on the test dataset. For an in-depth comparison between the results of our replications and the reported performance in these studies, we kindly invite the reader to consult the Appendix section \ref{sec:appendix} of this paper.



    \item \textbf{Intervention model:} we introduce GuiltBERT, a model based on the BERT architecture but finetuned on Guilt data. This model has 12 layers, 768 hidden units, and 12 attention heads. To fine-tune the model, random samples from the Vent dataset \cite{ventlykousas2019} were employed. In this dataset, samples from the guilt class served as positive samples, while samples from other classes (not-guilt classes) were considered negative samples. The model was trained with a maximum sequence length of 128 tokens and a batch size of 32 samples. The training process used the hyperparameters presented in Table \ref{tab:guiltbert_hyperparams}. We also employed an early-stopping mechanism to optimize loss and found the model achieved a validation loss of 1.853 after 8 epochs.

Table \ref{tab:guiltbert_loss} shows the training and validation loss of GuiltBERT for each epoch.

\begin{center}
\begin{table}[ht!]
\centering
\begin{tabular}{l|l}
\hline
\textbf{Hyperparameter} & \textbf{Value} \\ \hline

Optimizer & AdamW \\ 
Learning rate & 2e-5 \\
Batch size & 32 \\ 
Maximum sequence length & 128 \\ 
Weight decay rate & 0.01 \\ 
Learning rate warmup steps & 1,000 \\
Number of epochs & 8 \\ 
Early stopping &  loss \\ 

\hline
\end{tabular}
\caption{Training Hyperparameters of GuiltBERT}
\label{tab:guiltbert_hyperparams}
\end{table}
\end{center}

\begin{center}
\begin{table}[ht!]
\begin{tabular}{c|c|c}
\hline

Train Loss & Validation Loss & Epoch \\ \hline
2.0976 & 1.8593 & 0 \\ 
1.9643 & 1.8547 & 1 \\ 
1.9651 & 1.9003 & 2 \\ 
1.9608 & 1.8617 & 3 \\ 
1.9646 & 1.8756 & 4 \\ 
1.9626 & 1.9024 & 5 \\ 
1.9574 & 1.8421 & 6 \\ 
1.9594 & 1.8632 & 7 \\ 
1.9616 & 1.8530	& 8 \\
\hline

\end{tabular}
\caption{GuiltBERT finetuning Results}
\label{tab:guiltbert_loss}
\end{table}
\end{center}


\end{enumerate}

\subsection{The data processing}
To evaluate the performance of our language models on emotion detection we use the datasets of ISEAR and VIC and compare the results in a multiclass classification setting with the performance of other pre-trained transformer models which are the state-of-the-art models in emotion detection. Then, to showcase our model performance on guilt detection we also compare these models on a binarized version of VIC that splits the samples into guilt and no-guilt classes. 

The following items describe the processing we did over the emotion detection datasets used for evaluation on both tasks emotion detection and guilt detection. Table \ref{tab:data_stats} presents a summary of the statistics of the datasets.

\begin{table*}[h!]
\centering
\begin{tabular}{l ccc|ccc}
\hline
\multicolumn{1}{c}{} & \multicolumn{3}{c|}{Number of samples} & \multicolumn{3}{c}{Avg. words per text}
\\
\multicolumn{1}{c}{Emotion} 
& \multicolumn{1}{c}{ISEAR} & \multicolumn{1}{c}{VIC balanced} & \multicolumn{1}{c|}{VIC binary}
& \multicolumn{1}{c}{ISEAR} & \multicolumn{1}{c}{VIC balanced} & \multicolumn{1}{c}{VIC binary}
\\ \hline

guilt & 1066 & 167 & 2311 & 11.57 & 28.52 & 30 \\
disgust & 1073 & 167 & -- & 11.17 & 20.54 & -- \\
shame & 1066 & 167 & -- & 10.9 & 23.56 & -- \\
anger & 1083 & 167 & -- & 12.28 & 25.5 & -- \\
fear & 1078 & 167 & -- & 12.06 & 23.89 & -- \\
sadness & 1076 & 167 & -- & 9.93 & 19.96 & -- \\
joy & 1086 & 167 & -- & 9.81 & 18.72 & -- \\
other & -- & 167 & -- & -- & 30.29 & -- \\
no-guilt & -- & -- & 2311 & -- & -- & 26.3 \\
\hline
\end{tabular}
\caption{@ dataset comparison per class}
\label{tab:data_stats}
\end{table*}




\begin{enumerate}    
    
    \item \textbf{ISEAR:} In order to process ISEAR and compare our experiments with those reported in  \cite{adoma_etal_2020} we followed the steps described in their report including the removal of special characters, tags, irregular expressions, stopwords, and double spacing. We also removed examples resulting in sentences of single words, resulting in 7528 cleaned examples, 61 fewer samples than the cleaned version reported by \cite{adoma_etal_2020}.

    \item \textbf{VIC balanced:} To ensure consistent comparisons across datasets, we balanced the VIC dataset. This involved taking the intersection of VIC and ISEAR classes, along with introducing a new "other" class. Subsequently, we randomly selected 167 samples for each class, aligning with the minimum sample count in the dataset to ensure equal class sizes.
    
    \item \textbf{VIC binary:} 
    To perform the task of Guilt detection we binarized the original VIC dataset, resulting in equal parts guilt and no guilt samples as described in \citet{meque2023machine}
\end{enumerate}

Table \ref{tab:dataset_splits} outlines the sample distribution for the datasets' split. 
 
\begin{table}[ht!]
\centering
\begin{tabular}{l c|c|c|c}
\hline
\textbf{Dataset} & Train & Valid & Test & Total\\ \hline
ISEAR & 6029 & 746 & 753 & 7528\\ 
VIC balanced  & 1069 & 133 & 134 & 1336 \\ 
VIC Binary & 3701 & 458 & 463 & 4622\\ 
\hline
\end{tabular}
\caption{The dataset splits}
\label{tab:dataset_splits}
\end{table}

\section{Result}
In this section, we present the outcomes of our evaluations on three distinct datasets, showcasing the results of our proposed models for general emotion detection and guilt detection. Table \ref{tab:result-isear} and Table \ref{tab:result-vicbal} illustrate the results obtained for the emotion detection task using the ISEAR and the VIC balanced datasets respectively, while Table \ref{tab:result-vicbin} presents the results obtained for the guilt detection task using the Vic binary dataset.
To assess the performance of our models and provide meaningful comparisons we report the standard metrics of precision, recall, and F1 scores.

\begin{table*}[h!]
\centering
\begin{tabular}{l ccc|ccc|ccc}
\hline
\multicolumn{1}{c}{} & \multicolumn{3}{c|}{BERT} & \multicolumn{3}{c|}{RoBERTa} & \multicolumn{3}{c}{GuiltBERT}
\\
\multicolumn{1}{c}{ISEAR} 
& \multicolumn{1}{c}{P} & \multicolumn{1}{c}{R} & \multicolumn{1}{c|}{F1}
& \multicolumn{1}{c}{P} & \multicolumn{1}{c}{R} & \multicolumn{1}{c|}{F1}
& \multicolumn{1}{c}{P} & \multicolumn{1}{c}{R} & \multicolumn{1}{c}{F1}
\\ \hline

guilt & 0.55 & 0.37 & 0.44 & 0.63 & 0.55 & 0.59 & 0.44 & 0.54 & 0.49 \\
disgust & 0.76 & 0.81 & 0.78 & 0.65 & 0.71 & 0.68 & 0.75 & 0.82 & 0.78 \\
shame & 0.71 & 0.8 & 0.75 & 0.42 & 0.48 & 0.45 & 0.76 & 0.78 & 0.77 \\
anger & 0.67 & 0.57 & 0.62 & 0.61 & 0.39 & 0.47 & 0.65 & 0.6 & 0.62 \\
fear & 0.46 & 0.62 & 0.53 & 0.74 & 0.81 & 0.77 & 0.52 & 0.47 & 0.49 \\
sadness & 0.59 & 0.45 & 0.51 & 0.67 & 0.67 & 0.67 & 0.62 & 0.5 & 0.55 \\
joy & 0.54 & 0.65 & 0.59 & 0.73 & 0.85 & 0.78 & 0.6 & 0.64 & 0.62 \\
\hline
Macro (avg) & 0.61 & 0.61 & 0.60 & 0.64 & 0.64 & 0.63 & 0.62 & 0.62 & 0.62 \\
\hline
\end{tabular}
\caption{Models performance on emotion detection using ISEAR}
\label{tab:result-isear}
\end{table*}

\begin{table*}[h!]
\centering
\begin{tabular}{l ccc|ccc|ccc}
\hline
\multicolumn{1}{c}{} & \multicolumn{3}{c|}{BERT} & \multicolumn{3}{c|}{RoBERTa} & \multicolumn{3}{c}{GuiltBERT}
\\
\multicolumn{1}{c}{VIC balanced} 
& \multicolumn{1}{c}{P} & \multicolumn{1}{c}{R} & \multicolumn{1}{c|}{F1}
& \multicolumn{1}{c}{P} & \multicolumn{1}{c}{R} & \multicolumn{1}{c|}{F1}
& \multicolumn{1}{c}{P} & \multicolumn{1}{c}{R} & \multicolumn{1}{c}{F1}
\\ \hline

guilt & 0.75 & 0.75 & 0.75 & 0.86 & 0.76 & 0.81 & 0.64 & 0.75 & 0.69 \\
disgust & 0.83 & 0.76 & 0.79 & 0.88 & 0.93 & 0.9 & 0.81 & 0.68 & 0.74 \\
shame & 0.75 & 0.6 & 0.67 & 1 & 0.87 & 0.93 & 0.75 & 0.6 & 0.67 \\
anger & 0.87 & 0.81 & 0.84 & 0.74 & 0.85 & 0.79 & 0.87 & 0.81 & 0.84 \\
fear & 0.65 & 0.87 & 0.74 & 0.69 & 1 & 0.81 & 0.76 & 0.87 & 0.81 \\
sadness & 0.67 & 0.75 & 0.71 & 1 & 0.75 & 0.86 & 0.57 & 0.75 & 0.65 \\
joy & 0.92 & 0.69 & 0.79 & 1 & 0.94 & 0.97 & 0.75 & 0.75 & 0.75 \\
other & 0.86 & 1 & 0.93 & 0.76 & 0.81 & 0.79 & 0.94 & 0.89 & 0.92 \\
\hline
Macro (avg) & 0.79 & 0.78 & 0.78 & 0.87 & 0.86 & 0.86 & 0.76 & 0.76 & 0.76 \\
\hline
\end{tabular}
\caption{Models performance on emotion detection using VIC balanced}
\label{tab:result-vicbal}
\end{table*}

The results of our experiments for emotion detection showcase RoBERTa slightly outperforms the BERT and GuiltBERT models. Nonetheless, we remark the GuiltBERT outperforms the BERT and RoBERTa models in the guilt detection task 



\begin{table*}[h!]
\centering
\begin{tabular}{l ccc|ccc|ccc}
\hline
\multicolumn{1}{c}{} & \multicolumn{3}{c|}{BERT} & \multicolumn{3}{c|}{RoBERTa} & \multicolumn{3}{c}{GuiltBERT}
\\
\multicolumn{1}{c}{VIC binary} 
& \multicolumn{1}{c}{P} & \multicolumn{1}{c}{R} & \multicolumn{1}{c|}{F1}
& \multicolumn{1}{c}{P} & \multicolumn{1}{c}{R} & \multicolumn{1}{c|}{F1}
& \multicolumn{1}{c}{P} & \multicolumn{1}{c}{R} & \multicolumn{1}{c}{F1}
\\ \hline

guilt & 0.81 & 0.65 & 0.72 & 0.71 & 0.84 & 0.77 & 0.78 & 0.76 & 0.77 \\
no guilt & 0.7 & 0.84 & 0.76 & 0.82 & 0.68 & 0.74 & 0.76 & 0.77 & 0.77 \\
\hline
Macro (avg) & 0.76 & 0.75 & 0.74 & 0.77 & 0.76 & 0.76 & 0.77 & 0.77 & 0.77 \\
\hline
\end{tabular}
\caption{Models performance on guilt detection using VIC binary}
\label{tab:result-vicbin}
\end{table*}

To provide deeper insights into its performance, we present three confusion matrices showcasing GuiltBERT's classification results for each dataset. 
By analyzing these visual representations, we gain a comprehensive understanding of the GuiltBERT model performance in emotion classification, particularly in the context of guilt detection.

\begin{figure}[htp]
    \centering
    \includegraphics[width=8.2cm]{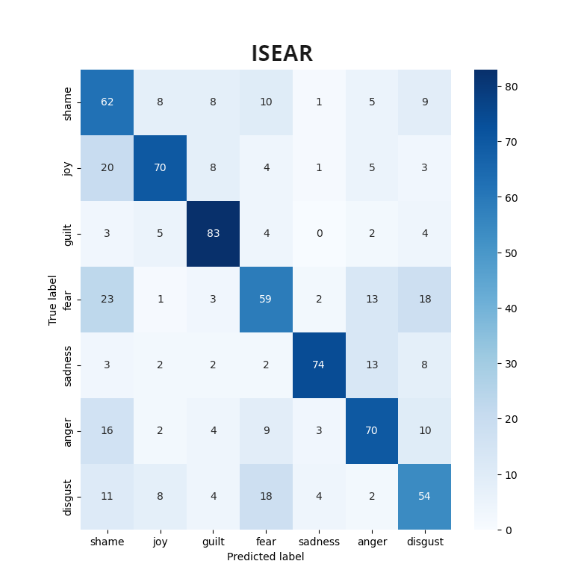}
    \caption{Confusion Matrix of GuiltBERT Model in the ISEAR Dataset}
\end{figure}

\begin{figure}[htp]
    \centering
    \includegraphics[width=8.2cm]{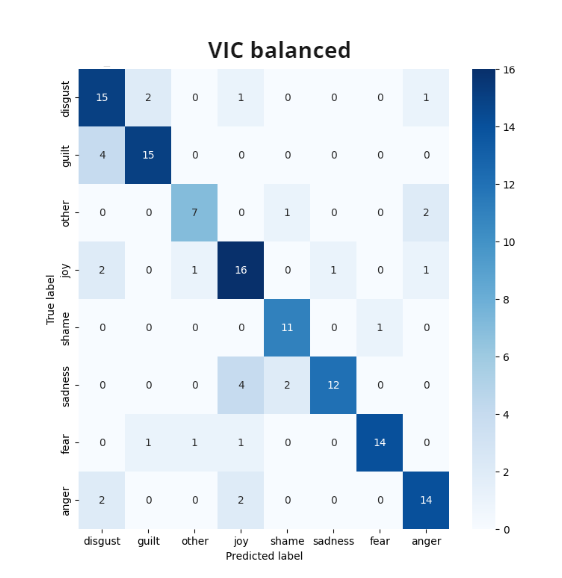}
    \caption{Confusion Matrix of GuiltBERT Model in the VIC balanced Dataset}    
\end{figure}

\begin{figure}[htp]
    \centering
    \includegraphics[width=8.2cm]{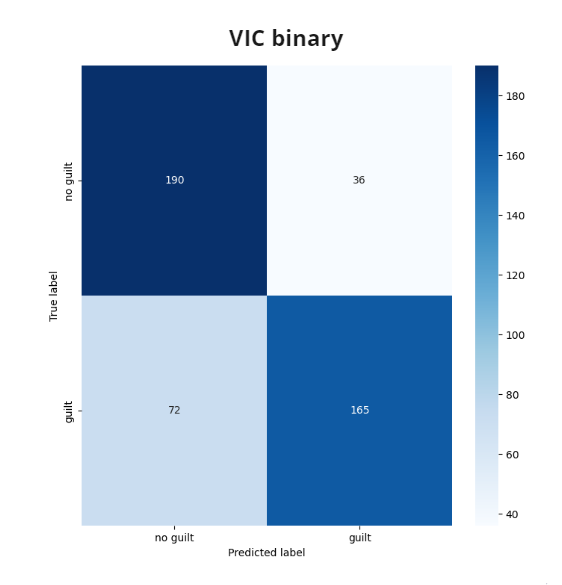}
    \caption{Confusion Matrix of GuiltBERT Model in the VIC binary Dataset}
\end{figure}

The findings reveal significant insights into GuiltBERT's capabilities for capturing the emotion of guilt. It demonstrates a remarkable accuracy in correctly identifying guilt, as evidenced by consistently high numbers of true positives across all experiments. This suggests that GuiltBERT excels in accurately detecting and classifying instances of guilt within the given datasets.

However, when it comes to emotion detection more broadly, GuiltBERT tends to occasionally confuse guilt with the emotions of disgust and shame. This can be attributed to the close relationship and overlapping characteristics between these negative emotions. While GuiltBERT's performance remains impressive, these instances of confusion highlight the complexity and intricacy of distinguishing between closely related emotions.


\section{Conclusion}
Researchers are developing increasingly sophisticated algorithms and models to identify, classify, and analyze emotional experiences in natural language text, and the study of emotions in NLP is a rapidly expanding field. 
However, despite the increasing interest in applying natural language processing techniques for emotion identification on social media, the emotion of guilt has received limited attention.


In this work, we explored the capability of language models in guilt detection. We also introduce GuiltBERT, a masked language model built upon BERT \cite{devlin2018bert}, as an emotion-specific approach to identifying guilt-related linguistic cues and patterns. 
This work makes significant contributions by shedding light on the detection of the less explored emotion of guilt and enabling further research in this domain. Our model achieves comparable results to existing emotion detection models while setting a new state-of-the-art for guilt detection.

As for future work, we plan to upgrade our GuiltBERT to be finetuned using Roberta and other architectures. 
Additionally, is worth exploring a new task called "explainable guilt" consisting of extracting the fragment of text implying a guilt emotion.






\section*{Acknowledgements}
The work was done with partial support from the Mexican Government through the grant A1-S-47854 of the CONACYT, Mexico, grants 20211784, 20211884, and 20211178 of the Secretaría de Investigación y Posgrado of the Instituto Politécnico Nacional, Mexico. The authors thank the CONACYT for the computing resources brought to them through the Plataforma de Aprendizaje Profundo para Tecnologías del Lenguaje of the Laboratorio de Supercómputo of the INAOE, Mexico, and acknowledge the support of Microsoft through the Microsoft Latin America Ph.D. Award.

\bibliography{custom}
\bibliographystyle{acl_natbib}

\appendix
\section{Training of model baselines} \label{sec:appendix}
The following describes the configuration we used to train BERT and RoBERTa models according to \citet{adoma_etal_2020} instructions. We preprocessed the ISEAR dataset by encoding the text inputs into numerical tokens using the pre-trained tokenizer of the selected model. Then set the maximum length of the input sequence to 200 tokens, and use a batch size of 16 for training the models. We use the SparseCategoricalCrossentropy loss function and Adam optimizer with a learning rate of 4e-5 to train the models for 10 epochs and use the validation dataset during training to monitor the model's performance and avoid overfitting.

\begin{table*}[h!]
\centering
\begin{tabular}{l ccc|ccc|ccc|ccc}
\hline
\multicolumn{1}{l}{} & \multicolumn{3}{c|}{BERT} & \multicolumn{3}{c|}{BERT\_adoma} & \multicolumn{3}{c|}{RoBERTa} & \multicolumn{3}{c}{RoBERTa\_adoma}
\\
\multicolumn{1}{l}{ISEAR} 
& \multicolumn{1}{c}{P} & \multicolumn{1}{c}{R} & \multicolumn{1}{c|}{F1}
& \multicolumn{1}{c}{P} & \multicolumn{1}{c}{R} & \multicolumn{1}{c|}{F1}
& \multicolumn{1}{c}{P} & \multicolumn{1}{c}{R} & \multicolumn{1}{c|}{F1}
& \multicolumn{1}{c}{P} & \multicolumn{1}{c}{R} & \multicolumn{1}{c}{F1}
\\ \hline

guilt & 0.55 & 0.37 & 0.44 & 0.65 & 0.69 & 0.67 & 0.63 & 0.55 & 0.59 & 0.62 & 0.76 & 0.68 \\
disgust & 0.76 & 0.81 & 0.78 & 0.71 & 0.63 & 0.67 & 0.65 & 0.71 & 0.68 & 0.76 & 0.69 & 0.73 \\
shame & 0.71 & 0.8 & 0.75 & 0.63 & 0.57 & 0.6 & 0.42 & 0.48 & 0.45 & 0.69 & 0.62 & 0.65 \\
anger & 0.67 & 0.57 & 0.62 & 0.56 & 0.57 & 0.57 & 0.61 & 0.39 & 0.47 & 0.67 & 0.59 & 0.62 \\
fear & 0.46 & 0.62 & 0.53 & 0.74 & 0.76 & 0.75 & 0.74 & 0.81 & 0.77 & 0.8 & 0.81 & 0.8 \\
sadness & 0.59 & 0.45 & 0.51 & 0.76 & 0.8 & 0.78 & 0.67 & 0.67 & 0.67 & 0.77 & 0.81 & 0.79 \\
joy & 0.54 & 0.65 & 0.59 & 0.84 & 0.91 & 0.88 & 0.73 & 0.85 & 0.78 & 0.9 & 0.96 & 0.93 \\
\hline
Macro (avg) & 0.61 & 0.61 & 0.60 & 0.70 & 0.70 & 0.70 & 0.64 & 0.64 & 0.63 & 0.74 & 0.75 & 0.74 \\
\hline
\end{tabular}
\caption{@ bert and roberta replica of Adoma and Adoma-original on ISEAR}
\label{tab:result-val}
\end{table*}


\end{document}